\documentclass[lettersize,journal]{IEEEtran}
\usepackage{amsmath,amsfonts}
\usepackage{algorithmic}
\usepackage{algorithm}
\usepackage{array}
\usepackage[caption=false,font=normalsize,labelfont=sf,textfont=sf]{subfig}
\usepackage{textcomp}
\usepackage{stfloats}
\usepackage{url}
\usepackage{verbatim}
\usepackage{booktabs}
\usepackage{pifont} 
\usepackage{graphicx}
\usepackage{cite}
\usepackage{multirow}
\usepackage{graphicx}  
\usepackage{amsmath}   
\usepackage{amssymb}   
\begin{document}

\title{RIE-SenseNet: Riemannian Manifold Embedding of Multi-Source Industrial Sensor Signals for Robust Pattern Recognition}

\author{Xu Wang, Puyu Han, Jiaju Kang, Weichao Pan, Luqi Gong
\thanks{Xu Wang and Weichao Pan are with the Shandong Jianzhu University.
}
\thanks{Puyu Han is with Southern University of Science and Technology.}
\thanks{Jiaju Kang, Puyu Han and Xu Wang are with the Fuxi AI Lab.}
\thanks{Luqi Gong is with the Zhejiang Lab, Nanhu Headquarters.}
\thanks{Corresponding author: Luqi Gong (luqi@zhejianglab.com)}
}



\maketitle

\begin{abstract}
Industrial sensor networks produce complex signals with nonlinear structure and shifting distributions. We propose RIE-SenseNet, a novel geometry-aware Transformer model that embeds sensor data in a Riemannian manifold to tackle these challenges. By leveraging hyperbolic geometry for sequence modeling and introducing a manifold-based augmentation technique, RIE-SenseNet preserves sensor signal structure and generates realistic synthetic samples. Experiments show RIE-SenseNet achieves \>90\% F1-score, far surpassing CNN and Transformer baselines. These results illustrate the benefit of combining non-Euclidean feature representations with geometry-consistent data augmentation for robust pattern recognition in industrial sensing.
\end{abstract}


\section{Introduction}

\IEEEPARstart{I}{ndustrial} sensor systems are widely applied in condition monitoring, fault detection, and quality control \cite{b1,b2,b3}. For example, in a pipeline intrusion detection deployment using distributed fiber-optic sensors \cite{b3}, multiple sensors must detect abnormal disturbance events amid significant environmental noise. The goal is to reliably classify such events in real time, which is very challenging. These time-series signals often pose two critical challenges:

1. Structural complexity — Sensor data exhibit complex, nonlinear dependencies challenging for Euclidean models\cite{b5}.

2. Domain instability — Environmental variations cause distribution shifts, hindering model generalization.

Euclidean-based CNNs or Transformers struggle with varying conditions \cite{b4}. Recent works have attempted to address these issues \cite{b6}. Zhai $et$ $al$.\cite{b6-1} used a multibranch deep network to mitigate sensor drift, but it is limited to gas sensing. Yuan $et$ $al$.\cite{b6-2} introduced a VAE-based multi-source domain adaptation using optimal transport, which reduces cross-domain discrepancy but requires complex alignment and still relies on effective representation learning\cite{b7}.

Existing solutions inadequately address complexity and domain shifts simultaneously.

To address these issues, we propose RIE-SenseNet, effectively modeling complex industrial sensor data. Its main contributions are:

\begin{itemize}
\item \textbf{Geometry-aware representation:} We embed multi-channel sensor sequences into a hyperbolic Riemannian manifold using Möbius operations, preserving latent hierarchical structures and improving feature expressiveness.
\item \textbf{Curvature-aware Transformer:} The backbone operates in non-Euclidean space with learnable curvature, enabling robust spatiotemporal modeling of irregular signals.
\item \textbf{Manifold-based augmentation:} Our augmentation pipeline perturbs embeddings and reconstructs realistic signals, enhancing generalization.
\item \textbf{Superior performance:} Experiments demonstrate RIE-SenseNet achieves \>90\% F1-score, greatly reducing false alarms compared to CNNs and Transformers.
\end{itemize}

RIE-SenseNet thus offers a generalizable and robust solution for sensor-based pattern recognition under complex and unstable conditions.

\section{Methods}

\subsection{Problem Context and Approach}
In industrial intrusion detection with Sensor Data, the goal is to classify various disturbance events from high-dimensional sensor signal data. We propose a two-pronged approach: (1) augment the Sensor Data dataset on a Riemannian manifold to address data scarcity and improve robustness, and (2) introduce a novel model, RIE-SenseNet, that operates in hyperbolic space for effective high-dimensional time-series classification. This approach provides a complete development process for industrial event classification models. Leveraging manifold augmentation and hyperbolic representations, RIE-SenseNet achieves superior real-world generalization.

\begin{figure}[htbp]
\centering
\includegraphics[width=0.85\linewidth]{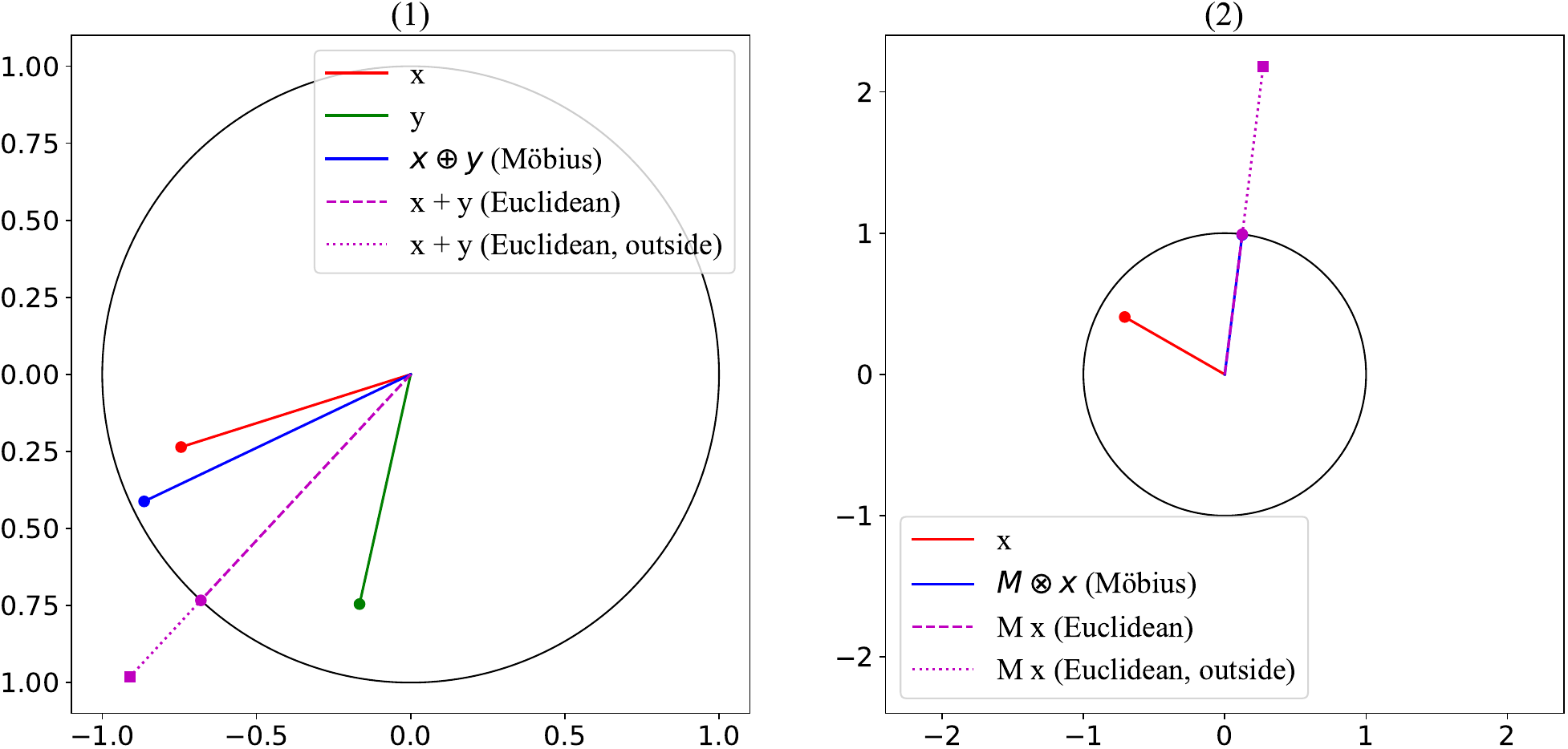}
\caption{Comparison of Möbius and Euclidean operations in the Poincaré disk model. The left subplot illustrates Möbius addition (blue) and Euclidean addition (magenta), showing that Möbius addition preserves hyperbolic geometry while Euclidean addition may lead to distortions outside the disk. The right subplot highlights the same behavior for Möbius matrix-vector multiplication versus Euclidean multiplication, with Möbius operations maintaining the integrity of the hyperbolic space.}
\label{fig:mobius}
\end{figure}

\begin{figure*}[htbp]
\centering
\includegraphics[width=0.6\linewidth]{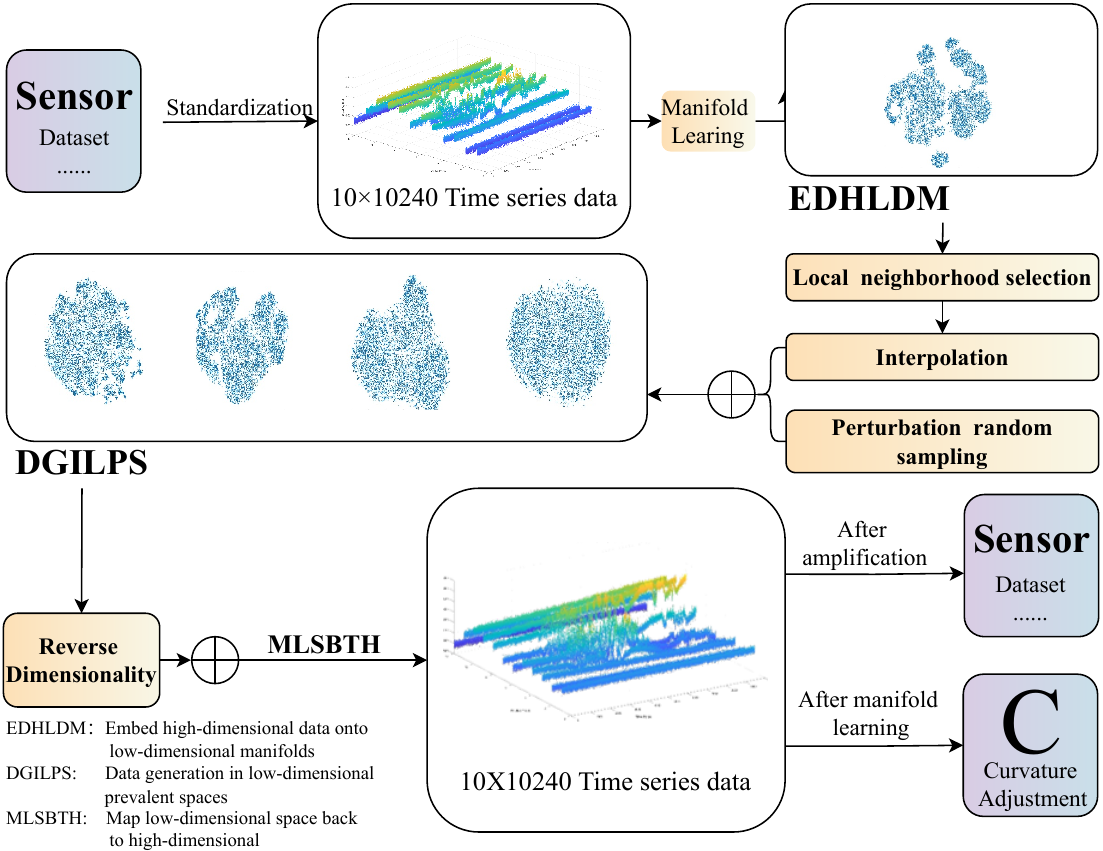}
\caption{Manifold-based data augmentation pipeline for Sensor Data signals.}
\label{fig:augmentation}
\end{figure*}

\subsection{Riemannian Manifold Data Augmentation}

We augment high-dimensional signals using a Riemannian manifold. Using invertible normalizing flows, original signals are mapped into a latent manifold space, where perturbations are applied to generate counterfactual examples. Signals are inversely mapped, preserving geometry and robustness. GAN-based flows approximate the manifold structure to generate realistic synthetic samples.

Specifically, we learn a diffeomorphic mapping via normalizing flows~\cite{b11,b15}, allowing realistic signal perturbations in the latent space~\cite{b12}. The augmentation respects hierarchical patterns typical in Sensor Data. This process produces a diverse and structurally faithful training set, effectively reducing enhancing model robustness~\cite{b10}.

\subsection{Hyperbolic Mapping and Model Architecture}

RIE-SenseNet integrates hyperbolic geometry into a transformer-based sequence classifier to better capture complex patterns in the data. First, we map input sensor signals into a hyperbolic space (Poincaré ball model) of suitable curvature. This is done via Möbius scalar multiplication (also viewed as hyperbolic linear transformation) defined as:
\begin{equation}
M(m,x,c) = \tanh\left( \frac{\|mx\|}{\|x\|} \cdot \text{atanh}(\sqrt{c} \cdot \|x\|) \right) \cdot \frac{mx}{\|mx\|},
\end{equation}where $x$ is the input feature vector, $m$ is a weight matrix, and $c$ is the curvature. This operation projects $x$ preserving geometric structure. We then fuse embedded signal features with any auxiliary contextual data using Möbius addition (the hyperbolic analog of vector addition). Möbius addition $(x \oplus y)$ for vectors $x,y$ (with curvature $c$) is defined as:
\begin{equation}
A(x, y, c) = \frac{(1 + 2c \langle x, y \rangle + c \|y\|^2)x + (1 - c \|x\|^2)y}{1 + 2c \langle x, y \rangle + c^2 \|x\|^2 \|y\|^2}.
\end{equation}which ensures the result stays in the curved space. These operations embed representations on a manifold, better reflecting data structure than Euclidean geometry. Figure~\ref{fig:mobius} shows Möbius operations keep points within the disk, unlike Euclidean ones which distort geometry. In other words, Möbius transformations preserve hyperbolic geometry, whereas Euclidean additions or multiplications do not. RIE-SenseNet maintains relational fidelity crucial for high-dimensional signals \cite{b6-4}. 


\begin{table*}[ht]
\centering
\caption{Classification Performance by Event Class.}
\label{tab:cls_perf}
\begin{tabular}{lccccccccccccccccccc}
\toprule
\textbf{Model} & \multicolumn{3}{c}{Impact Hammer} & \multicolumn{3}{c}{Knocking} & \multicolumn{3}{c}{Percussion Drill} \\
& Precision & Recall & F1-score & Precision & Recall & F1-score & Precision & Recall & F1-score \\
\midrule
SVM & 39.2 & 38.4 & 38.7 & 34.7 & 34.2 & 34.5 & 36.5 & 36.0 & 36.2 \\
Random Forest & 61.8 & 60.5 & 61.2 & 59.3 & 58.8 & 59.0 & 62.2 & 61.8 & 62.0 \\
CNN & 63.8 & 62.4 & 63.1 & 61.7 & 61.2 & 61.5 & 64.1 & 63.6 & 63.8 \\
Transformer & 63.7 & 62.9 & 63.3 & 62.1 & 61.5 & 61.8 & 64.0 & 63.5 & 63.7 \\
\textbf{RIE-SenseNet} & \textbf{92.4} & \textbf{91.9} & \textbf{92.1} & \textbf{91.7} & \textbf{91.3} & \textbf{91.6} & \textbf{93.2} & \textbf{92.7} & \textbf{92.9} \\
\toprule
\textbf{Model} & \multicolumn{3}{c}{Large Excavator} & \multicolumn{3}{c}{Subway} & \multicolumn{3}{c}{Engine Idle} \\
& Precision & Recall & F1-score & Precision & Recall & F1-score & Precision & Recall & F1-score \\
\midrule
SVM & 31.8 & 31.2 & 31.4 & 33.3 & 32.9 & 33.1 & 35.8 & 35.3 & 35.6 \\
Random Forest & 57.8 & 57.2 & 57.5 & 59.1 & 58.6 & 58.8 & 60.7 & 60.2 & 60.5 \\
CNN & 60.3 & 59.8 & 60.0 & 62.0 & 61.5 & 61.8 & 63.2 & 62.8 & 63.0 \\
Transformer & 61.4 & 60.8 & 61.0 & 63.0 & 62.5 & 62.8 & 63.8 & 63.3 & 63.5 \\
\textbf{RIE-SenseNet} & \textbf{90.6} & \textbf{90.1} & \textbf{90.3} & \textbf{91.9} & \textbf{91.6} & \textbf{91.8} & \textbf{92.2} & \textbf{91.8} & \textbf{92.0} \\
\bottomrule
\end{tabular}
\end{table*}

\subsection{Hyperbolic Attention Mechanism}
RIE-SenseNet employs a transformer-style attention, modified to operate in hyperbolic space. Queries and keys are first compared via the standard inner-product to compute a similarity score:
\begin{equation}
C = \frac{q \cdot k^T}{\|q\| \cdot \|k\|},
\end{equation}
We then map this similarity into hyperbolic space by interpreting it as a point on the hyperbolic curve: specifically, we compute an intermediate value:
\begin{equation}
H_{input} = 1 + c(C - 1),
\end{equation} which lies on the domain of the hyperbolic arccosine. The attention distance in hyperbolic space is obtained as:

\begin{equation}
H = \text{arcosh}(\max(H_{input}, 1.0)),
\end{equation}Here, $\mathrm{arcosh}$ converts the cosine similarity into a hyperbolic distance measure, and we clamp the minimum to 1.0 to ensure the argument is in the valid range for arcosh.

These hyperbolic distances $H$ are converted to attention weights via softmax.:

\begin{equation}
W = \text{softmax}(-H).
\end{equation}

This operation projects $x$ into the hyperbolic space while preserving its geometric structure, effectively acting as a Poincaré mapping. In contrast, a standard linear mapping $m x$ in Euclidean space does not guarantee the result stays on the manifold or maintains hierarchical relationships. By using the Poincaré ball model (a hyperbolic space of constant negative curvature $c$), we ensure that combined features remain inside a bounded radius (the unit disk for $c=1$). Curvature $c$ is a learnable parameter adapting geometry during training. In summary, the Möbius transformation enables sensor features to be embedded in a non-Euclidean space that preserves their inherent structure far better than a Euclidean projection.

\section{Experiments}

\subsection{Datasets and Preprocessing}

We evaluated RIE-SenseNet on two Sensor Data datasets. The first is a public dataset from Beijing Jiaotong University~\cite{b17}, consisting of 15,419 labeled samples across six intrusion event types: background noise, digging, knocking, shaking, watering, and walking. To enhance generalization, we applied our manifold-based GAN augmentation: each category was expanded with 20,000 synthetic samples. As shown in Figure~\ref{fig:augmentation}, raw 10$\times$10240 time-series signals are embedded into a low-dimensional manifold (EDHLDM), where data augmentation is performed via local sampling, interpolation~\cite{b9}, and perturbation (DGILPS), followed by mapping back with curvature adjustment (MLSBTH), thus preserving the geometric structure.

The second dataset is an industrial dataset we collected, covering six real-world event types: impact hammer, knocking, percussion drill, large excavator, subway passage, and engine idle. Each category includes 1,500 samples recorded via fiber-optic sensing. The backscattered signals were converted via photodetector and digitized at 10MS/s. All data were standardized to normalize feature scales.

\subsection{Experimental Setup}

Models used an 80/20 train-test split on one NVIDIA A100 GPU. Classification metrics include \textit{Accuracy, Precision, Recall} and  \textit{F1-score}. We conduct ablation studies and hyperparameter sensitivity tests for manifold curvature and hyperbolic attention.

\subsection{Baselines}
We conducted all experiments on a single NVIDIA A100 GPU for consistency in computational conditions. The performance of RIE-SenseNet was compared against several baseline methods. These baselines include: 
\begin{itemize}
\item \textbf{Traditional Sensor Data classifiers using classical machine learning} – specifically, a Support Vector Machine (SVM) and a Random Forest – trained on the raw time-domain signals

\item  \textbf{Deep learning models} – an end-to-end convolutional neural network (CNN) and a Transformer-based model – trained on the same spatio-temporal data (without hyperbolic mapping)

\item  \textbf{Generative augmentation models}, to evaluate the impact of our data augmentation, where we compare training with the manifold GAN-augmented data versus training with standard data augmentation techniques
\end{itemize}
All models (including baselines and RIE-SenseNet) are evaluated using the common classification metrics of Accuracy, Precision, Recall, and F1 Score. Metrics were computed per event and overall; ablation studies evaluated key components.

\subsection{Quantitative Results}
Table~\ref{tab:cls_perf} summarizes classification results on the industrial dataset. Looking at the baseline results, we observe that classical machine learning models (SVM, Random Forest) achieve significantly lower accuracy (F1 in the 30–60\% range) compared to deep learning models (CNN, Transformer around 63\% F1). Even within deep models, the CNN and the standard Transformer perform similarly (~63\% F1), indicating that simply increasing model complexity (from CNN to Transformer) yields little improvement under these conditions. RIE-SenseNet outperforms all of these baselines by a large margin, demonstrating that neither traditional ML nor standard deep architectures can adequately handle the structural complexity and domain shifts present in this dataset.

\begin{table}[h]
\centering
\caption{Performance Comparison of RIE-SenseNet with Baseline Models}
\label{tab:ablation}
\renewcommand{\arraystretch}{1.2} 
\resizebox{1\linewidth}{!}{
\begin{tabular}{c|c|ccccc}
\hline
\multirow{2}{*}{\textbf{Method}} & \multirow{2}{*}{\textbf{Augmented}} & \textbf{Accuracy} & \textbf{Precision} & \textbf{Recall} & \textbf{F1 Score} \\
 & & (\%) & (\%) & (\%) & (\%) \\
\hline
\multirow{2}{*}{SVM} & - & 85.3 & 84.7 & 85.1 & 84.9 \\
 & \checkmark & 87.2 & 86.0 & 87.0 & 86.5 \\
\multirow{2}{*}{Random Forest} & - & 87.6 & 86.4 & 87.1 & 86.7 \\
 & \checkmark & 89.1 & 87.9 & 88.7 & 88.3 \\
\multirow{2}{*}{CNN} & - & 88.4 & 87.2 & 88.0 & 87.6 \\
 & \checkmark & 90.0 & 88.8 & 89.6 & 89.2 \\
Transformer & - & 90.0 & 89.4 & 89.8 & 89.6 \\
(Spatio-temporal)& \checkmark & 91.5 & 90.2 & 91.1 & 90.7 \\
\hline
\multirow{2}{*}{\textbf{RIE-SenseNet}} & - & 92.4 & 91.8 & 92.2 & 92.0 \\
 & \checkmark & \textbf{93.0} & \textbf{92.5} & \textbf{93.0} & \textbf{92.8} \\
\hline
\end{tabular}
}
\end{table}

Table~\ref{tab:ablation} presents overall performance with/without augmentation. RIE-SenseNet achieves the highest accuracy of 93.0\% with augmentation.

\subsection{Analysis and Discussion}
The results confirm the superiority of RIE-SenseNet, particularly in capturing long-range temporal dependencies and geometric relationships. RIE-SenseNet's margin highlights hyperbolic attention’s advantage. Moreover, while all models benefited from augmentation, RIE-SenseNet gained the most, indicating stronger synergy with Riemannian data diversity.

Ablation shows that removing hyperbolic geometry reduces F1 by $\sim$2.5\%. Similarly, replacing manifold augmentation with vanilla GANs lowers performance, suggesting that curvature-aligned perturbations are key to robustness.

RIE-SenseNet maintains stable metrics across classes without overfitting, ideal for real-time urban, transport, and pipeline monitoring \cite{b6-3}.

Removing hyperbolic geometry or manifold augmentation reduces F1 by 2.5\% and 1\%, respectively, and their combination explains the 30\% gain over baselines.

We further compared RIE-SenseNet with a representative manifold learning baseline. In this experiment, we first applied a classical manifold reduction (Isomap) to project the multi-sensor data into a 10-dimensional space, then trained a classifier on these features. This two-step approach yielded substantially lower accuracy (overall F1 below 70\%) compared to RIE-SenseNet. The result suggests that standard manifold learning alone — without an integrated sequence model — is insufficient to capture the complex patterns in the sensor signals. Moreover, existing hyperbolic embedding methods like Nickel et al. [10] are unsupervised and designed for static hierarchies, thus not directly applicable to our time-series classification task. In contrast, RIE-SenseNet jointly learns the manifold representation and temporal dynamics, which leads to its superior performance over these alternatives.

\section{Conclusion}

We have presented RIE-SenseNet, a Transformer-based framework for robust pattern recognition in industrial multi-sensor signal data. Hyperbolic embedding via Möbius operations preserves relational structures. Combined with our Riemannian manifold augmentation, this enables the generation of high-fidelity synthetic data and yields large performance gains. Experiments demonstrate RIE-SenseNet achieves state-of-the-art accuracy, reducing false alarms. Results underscore manifold techniques' utility for IIoT analytics. Future work includes adaptive curvature and broader sensor applications. As reflected in prior work, mapping data to a Riemannian manifold proves to be an effective strategy for improving generalization in complex sensing environments.

\end{document}